\documentclass{article}
\usepackage{spconf,amsmath,graphicx}

\usepackage{soul}
\usepackage{color,xcolor}
\usepackage{algorithm}
\usepackage{algorithmic}
\sethlcolor{yellow}
\usepackage{amsmath}
\usepackage{amsfonts}
\usepackage{subcaption}
\title{Version age-based client scheduling policy for federated learning}
\name{Xinyi Hu$^\dagger$\thanks{The work of X.~Hu and H.~H.~Yang was supported in part by the National Natural Science Foundation of China under Grant 62201504, in part by the Zhejiang Provincial Natural Science Foundation of China under Grant LGJ22F010001, and in part by the Zhejiang – Singapore Innovation and AI Joint Research Lab. The work of N. Pappas has been supported in part by the Swedish Research Council (VR), ELLIIT, the European Union (ETHER, 101096526), the European Union's Horizon Europe research and innovation programme under the Marie Skłodowska-Curie Grant Agreement No 101131481 (SOVEREIGN), and the Horizon Europe/JU SNS project ROBUST-6G (Grant Agreement no. 101139068).
} \qquad Nikolaos Pappas$^\mathsection$
\qquad Howard H. Yang$^\dagger$*}
\address{$^\dagger$ ZJU-UIUC Institute, Zhejiang University, China\\
$^\mathsection$ Department of Computer and Information Science, Link{\"o}ping University, Sweden \\
Email: xinyih@zju.edu.cn, nikolaos.pappas@liu.se, haoyang@intl.zju.edu.cn
}
\begin{document}
\maketitle
\begin{abstract}
Federated Learning (FL) has emerged as a privacy-preserving machine learning paradigm facilitating collaborative training across multiple clients without sharing local data. Despite advancements in edge device capabilities, communication bottlenecks present challenges in aggregating a large number of clients; only a portion of the clients can update their parameters upon each global aggregation. 
This phenomenon introduces the critical challenge of stragglers in FL and the profound impact of client scheduling policies on global model convergence and stability. 
Existing scheduling strategies address staleness but predominantly focus on either timeliness or content.
Motivated by this, we introduce the novel concept of Version Age of Information (VAoI) to FL. Unlike traditional Age of Information metrics, VAoI considers both timeliness and content staleness. Each client's version age is updated discretely, indicating the freshness of information. VAoI is incorporated into the client scheduling policy to minimize the average VAoI, mitigating the impact of outdated local updates and enhancing the stability of FL systems.
\end{abstract}
\section{Introduction}

Federated Learning (FL)~\cite{mcmahan2017communication,li2020federated,ZhaFenYan:20} stands as a privacy-preserving machine learning paradigm, facilitating collaborative training of a global model across multiple clients without necessitating the sharing of their local data. 
In contrast to traditional centralized machine learning approaches, FL conducts the training process on edge devices, with the exchange of intermediate parameters, such as weights or gradients, occurring between the central server and the clients.
Despite the increasing computational capabilities of edge devices that allow the deployment of large Deep Neural Networks, the communication bottleneck, characterized by limited bandwidth, imposes constraints on the number of clients eligible for aggregation in each communication round. 

The challenge posed by stragglers in FL has become increasingly critical compared to traditional data center training.
The efficacy of client scheduling policy significantly influences the convergence and stability of the global model~\cite{YanLiuQue:20,xia2020multi}, particularly in scenarios involving a large number of clients. 
Consequently, a plethora of scheduling strategies have been explored in FL to address the aforementioned challenge~\cite{ha2019coded, khanvalue}.
For example, \cite{xia2020multi} leverages a multi-armed bandit approach to determine decision outcomes for the next round of client scheduling based on feedback metrics such as communication and computation times associated with each action.
On the other hand, \cite{yang2020age} proposes a scheduling policy in the context of FL that introduces the concept of age of information (AoI) in FL.
The objective is to minimize the age of updates during each communication round.
Additional scheduling criteria include update significance, measured by model variance~\cite{kamp2019efficient} and gradient variance~\cite{chen2018lag}.
Despite the positive outcomes demonstrated by these approaches, they predominantly focus on addressing staleness in one aspect, either timeliness or content.

As pointed out by~\cite{yates2021age, buyukates2022version}, the staleness in both timeliness and content significantly impacts the convergence rate of networks with a source node and a set of receiver nodes.
Consequently, there is a justifiable expectation for a scheduling algorithm that transcends the consideration of timeliness alone and jointly incorporates content.
Unlike AoI, which measures the time elapsed, a novel age metric known as version age has recently emerged in the literature~\cite{yates2021age, abolhassani2021fresh, delfani2023version}. 
In the context of version age, each update at the source is treated as a version change, and the version age quantifies how many versions the information at the monitor is behind, compared to the version at the source. 
In contrast to the continuous nature of the original age metric, version age exhibits discrete steps. 
Specifically, the version age of a monitor increases by one when the source generates a newer version, indicating fresher information. 
During periods between version changes at the source, the version age of the monitor remains constant, signifying that the monitor still possesses the most recent information.

This paper introduces the concept of Version Age of Information (VAoI) to FL, aiming to address both the timeliness and content staleness aspects.
More precisely, when a client's local update significantly deviates from the latest global model, its version age is considered to have changed. 
A higher version age signifies decreased timeliness and content staleness jointly. The version age is updated at each communication round, and the scheduling policy is designed to minimize the average version age of the system.

\begin{figure}[t]
\centering\includegraphics[width=0.5\textwidth]{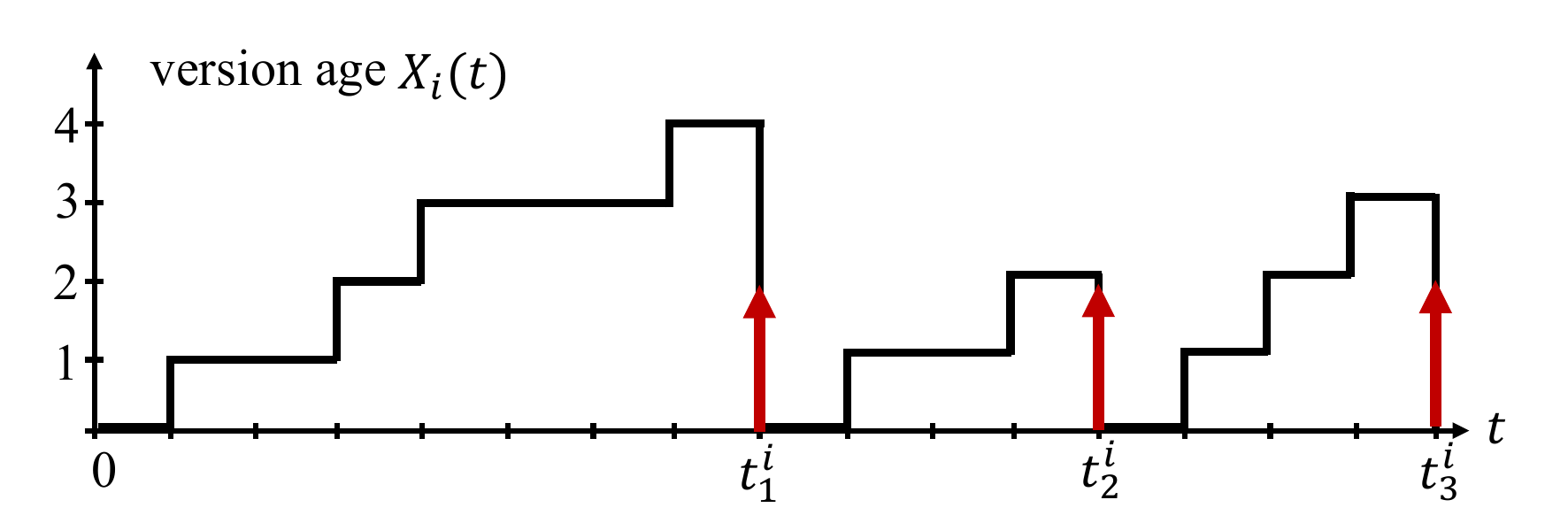} 
\caption{An example of the version age evolution over time, the upward arrows represent updates received by the client.}
\label{fig:version age}
\end{figure}

\section{System model}
\subsection{Federated Learning}
Consider a Federated Learning (FL) system consisting of a server and $N$ clients, in which client $i$ owns a loss function $f_i:\mathbb{R}^d\rightarrow\mathbb{R}$ constructed from its local dataset $\mathcal{D}_i$. 
The objective of all the participating entities in this system is to find a global model $\boldsymbol{w}\in \mathbb{R}^d$ that solves the problem 
\begin{equation}
    \label{equ:obj_fl}
    \min_{ \boldsymbol{w} } f( \boldsymbol{w} )=\sum_{i=1}^N \frac{n_i}{n} f_i( \boldsymbol{w} ),
\end{equation}
where $n_i=\vert\mathcal{D}_i\vert$ denotes the number of local data samples at client $i$ and $n=\sum_{j=1}^N\vert\mathcal{D}_j\vert$ is the total number of training samples across the system. 
In a typical communication round $t$, clients conduct local training based on the latest global model weights $\boldsymbol{w}_g^{t}$ broadcast by the server. 
Let $\boldsymbol{w}_i^t$ denote client $i$'s model weights after local training. 
At the end of round $t$, the server would collect local models from clients to update the global model via Federated Averaging (FedAvg)~\cite{mcmahan2017communication},
i.e., $\boldsymbol{w}_g^{t+1}= \sum_{i =1}^N \frac{n_i}{n} \boldsymbol{w}_i^t$.
Nevertheless, due to bandwidth constraints, the server can only select a subset of clients to participate in the model aggregation in each communication round as follows
\begin{equation}
    \boldsymbol{w}_g^{t+1}= \sum_{i \in \mathcal{S}_t} \beta_i^t \boldsymbol{w}_i^t,
\end{equation}
in which $\beta_i^t= |\mathcal{D}_i|/\sum_{j \in \mathcal{S}_t}|\mathcal{D}_j|$ represents the ratio of the local data samples in client $i$ over the total number of data samples in the selected subset $S_t$.
Previous research~\cite{chen2018lag,yang2020age} has shown that the choice of selection can largely affect the convergence rate of FL. 
In this regard, there is an urgent need for appropriate scheduling algorithms.

\subsection{Version Age of Information}

In this work, we draw on the concept of VAoI~\cite{yates2021age} as a metric for evaluating the freshness of updates. 
Recognizing that information freshness is intricately tied to both timeliness and content staleness, VAoI assesses the freshness of information for each node by monitoring version updates.
The source node consistently maintains the current (fresh) version of its status, ensuring that the source node always has a version age $\mathit{X}(t) = 0$. Commencing at time $t = 0$, status updates at the source node occur as a Poisson process denoted by $N(t)$. At any time $t > 0$, the most recent update at the source corresponds to version $N(t)$. If the current update at node $i$ is of version $N_i(t)$, then the version age at node $i$ is given by 
\begin{equation}
    X_i(t) = N(t) - N_i(t).
\end{equation}
Importantly, time and information freshness do not exhibit a linear correlation, as state updates on the source node do not occur at fixed intervals. It is during a state update on the source node that the information on nodes not receiving that update becomes more stale.
A sample trajectory of the version age, as depicted in Figure~\ref{fig:version age}, underscores its dependence on node-retained updates, demonstrating an irregular increase over time. Upon receipt of a new update, the version age is promptly reset to zero.

\subsection{Version Age-based Scheduling Policy}
\label{sec:VAS}
\begin{figure}[t]
\centering\includegraphics[width=0.5\textwidth]{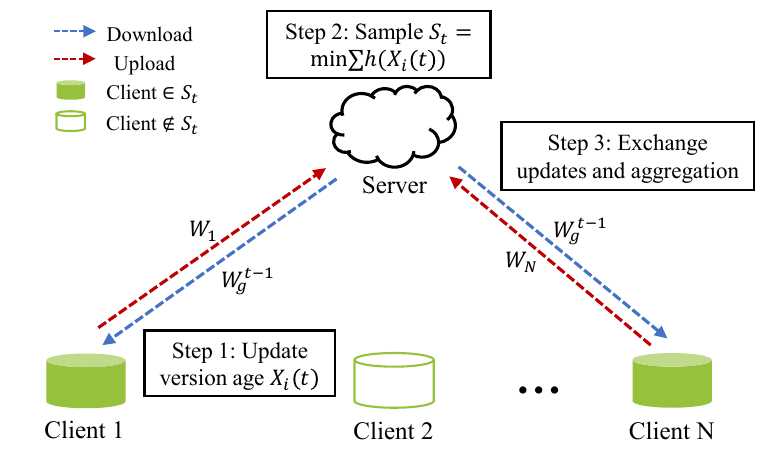} 
\caption{Overview of proposed scheduling policy.}
\label{fig:fl}
\end{figure}

Using the notion of VAoI, we devise our scheduling protocol as Figure~\ref{fig:fl} depicted.
The server initiates version updates in each communication round, where the progression of version age is contingent upon the $L_k$ norm distance, which is commonly used in high dimensional spaces, between the client's most recent update, denoted as $\boldsymbol{w}_i$, and the current global (fresh) model $\boldsymbol{w}_g^{t}$. 
Previous research \cite{aggarwal2001surprising} has demonstrated that the importance of the $L_k$ norm distance deteriorates rapidly with increasing dimensionality, especially for higher values of $k$. In the setup considered in this work, opting for lower $k$ might be preferable.
Therefore, we consider the Manhattan distance ($k=1$) \footnote{We can consider other types of distance by simply modifying the distance metric presented in our Algorithm without affecting its generality.}.

In instances where the calculated distance surpasses a predefined threshold $\tau$, the version age undergoes an increment. 
Furthermore, upon the client's selection, its version age is reset to zero.
For a generic client $i$, its version age evolves as articulated as follows
\begin{equation}
    \mathit{X}_i(t+1) = 
    \begin{cases}
    (\mathit{X}_i(t) + 1)(1-S(i)), &  \Vert \boldsymbol{w}_i - \boldsymbol{w}_g^{t} \Vert_1 \geq \tau,\\
    \mathit{X}_i(t)(1-S(i)), & \Vert \boldsymbol{w}_i - \boldsymbol{w}_g^{t} \Vert_1 < \tau,
    \end{cases}
\end{equation}
where $\mathit{X}_i(0) = 0$, and $S(i) \in \{0,1\}$ takes a value of 1 if the server selects client $i$ to participate in the aggregation process during communication round $t$. 
The magnitude of $\mathit{X}_i(t)$ directly correlates with the imperative for client update, thereby elevating the likelihood of its selection. 
Leveraging the concept of version age, we've crafted our scheduling protocol to ensure the server maintains the updates' freshness to the fullest extent.
This is achieved by minimizing the cost functions related to the version age, as illustrated below
\begin{equation}
    \label{equ:obj}
    \min \sum_{i=1}^N h(\mathit{X}_i(t)),
\end{equation}
here, the function $h(\cdot)$ symbolizes the sensitivity of the server to the version age of the updates. 
As discussed in ~\cite{kosta2017age,kosta2020cost}, performance degradation due to information aging may not be a linear function of time, even though the AoI grows at a unit rate. 
For example, consider the state estimation problem for a Gaussian linear time-invariant (LTI) system: if the system is stable, the state estimation error is a sublinear function of the AoI, converging to a finite constant~\cite{ornee2021sampling}; if the system is unstable, the state estimation error grows exponentially with the AoI~\cite{klugel2019aoi}. 
Whereas the version age does not grow at a unit rate, as Figure~\ref{fig:version age} shows, the mapping between system performance and VAoI is more likely to be non-linear. 
Hence, in this paper, we take $h(\cdot) = \exp(\cdot)$.

\subsection{Algorithm}

Here, we address the problem defined by Equation~\ref{equ:obj}. Initially, we assign distinct probabilities to clients based on their version ages. The probability $p_i(t)$, indicating the selection probability of client $i$ during communication round $t$, is determined through normalization, as follows
\begin{equation}
    \label{equ: prob}
    p_i = \frac{h(\mathit{X}_i(t))}{\sum_{j=1}^N h(\mathit{X}_j(t))}.
\end{equation}
Subsequently, the set of clients $S_t$ participating in the update for each communication round is sampled based on the probabilities $\{p_i\}$. This approach increases the likelihood of selecting stale clients. The specifics of forming the $S_t$ are outlined in Algorithm~\ref{alg:vaoi}. 
Finally, we present the FL process to solve Equation~\ref{equ:obj_fl}, as depicted in Algorithm~\ref{alg:fl}. 

\begin{algorithm}[h]
\caption{Federated Learning}
\label{alg:fl}

\begin{algorithmic}[1] 
\STATE \textbf{Initialize}: $\boldsymbol{w}_g^1, \{\boldsymbol{w}_i\}$ 
\FOR{$t = 1,2,...,T$ }
\STATE Invoke Alg.~\ref{alg:vaoi} with the input $\{X_i(t)\}, \{\boldsymbol{w}_i\}$ and $\boldsymbol{w}_g^{t}$ to obtain $S_t$ and update the version ages to $\{X_i(t+1)\}$
\FOR{each client $i \in \mathcal{S}_t$ in parallel}
\STATE Server send $\boldsymbol{w}_g^{t-1}$ to client $i$
\STATE Client $i$ updates $\boldsymbol{w}_i = \boldsymbol{w}_g^{t} - \eta \nabla f_i(\boldsymbol{w}_g^{t})$
\STATE Client $i$ uploads $\boldsymbol{w}_i$
\ENDFOR
\STATE Sever update global model $\boldsymbol{w}_g^{t+1} = \sum_{i\in S_t}\beta_i^t \boldsymbol{w}_i$

\ENDFOR
\end{algorithmic}
\end{algorithm}

\begin{algorithm}[h]
\caption{Version Age-based Scheduling}
\label{alg:vaoi}

\begin{algorithmic}[1] 
\STATE \textbf{Input}: $\{\mathit{X}_i(t)\}, \{\boldsymbol{w}_i\}$ and $\boldsymbol{w}_g^{t}$ 
\STATE \textbf{Initialize}: $S(i) = 0, i \in \{1,2,\dots,N\}$ 
\STATE Compute probabilities $\{p_i\}= \{\frac{\mathit{X}_i(t)}{\sum_{i=1}^N\mathit{X}_j(t)}\}$ 
\STATE Sample clients based on $\{p_i\}$ to obtain the set $S_t$
\FOR{each client $i =1,2,\dots, N$ in parallel}
\IF {client $i \in S_t $}
\STATE Assign $S(i) = 1$
\ENDIF
\STATE Update the version age: \\
\begin{small}
\begin{equation}
\setlength\abovedisplayskip{-0.2cm}
    \mathit{X}_i(t+1) = 
    \begin{cases}
    (\mathit{X}_i(t) + 1)(1-S(i)), &  \Vert \boldsymbol{w}_i - \boldsymbol{w}_g^{t} \Vert_1 \geq \tau,\\
    \mathit{X}_i(t)(1-S(i)), & \Vert \boldsymbol{w}_i - \boldsymbol{w}_g^{t} \Vert_1 < \tau
    \end{cases}
    \nonumber
\end{equation}
\end{small}

\ENDFOR
\STATE \textbf{Ouput}: $S_t, \{\mathit{X}_i(t+1)\}$ 
\end{algorithmic}
\end{algorithm}

It is crucial to note that in conventional FedAvg, every client has an equal probability of being selected, potentially resulting in some clients possessing outdated information due to not being updated for an extended period. 
Opting for such clients in the aggregation process may result in system instability due to the substantial disparities between their local models and the latest global model. In extreme instances, it may trigger catastrophic training failures~\cite{charles2021large, yang2022taming}.

\section{Numerical Results}
\label{sec:result}

In this section, we conduct simulations to validate the robustness and effectiveness of the 
Version Age-based Scheduling (VAS) policy.
The simulations utilize ResNet-18~\cite{he2016deep} applied to the CIFAR-100~\cite{krizhevsky2009learning} dataset. 
To introduce data heterogeneity, we distribute data with imbalanced labels among $100$ clients using a Dirichlet distribution~\cite{hsu2019measuring}. 
All additional hyperparameters in the simulation adhere to standard settings. Specifically, the parameter $\rho$, which governs the extent of class imbalance, is set to a value of 0.3. In each communication round, 10\% of clients are chosen to participate in aggregation, and upon selection, a client performs five local updates and then uploads. All results depicted in the figures represent averages over three trial simulation runs.

\begin{figure}[h]
\centering\includegraphics[width=0.5\textwidth]{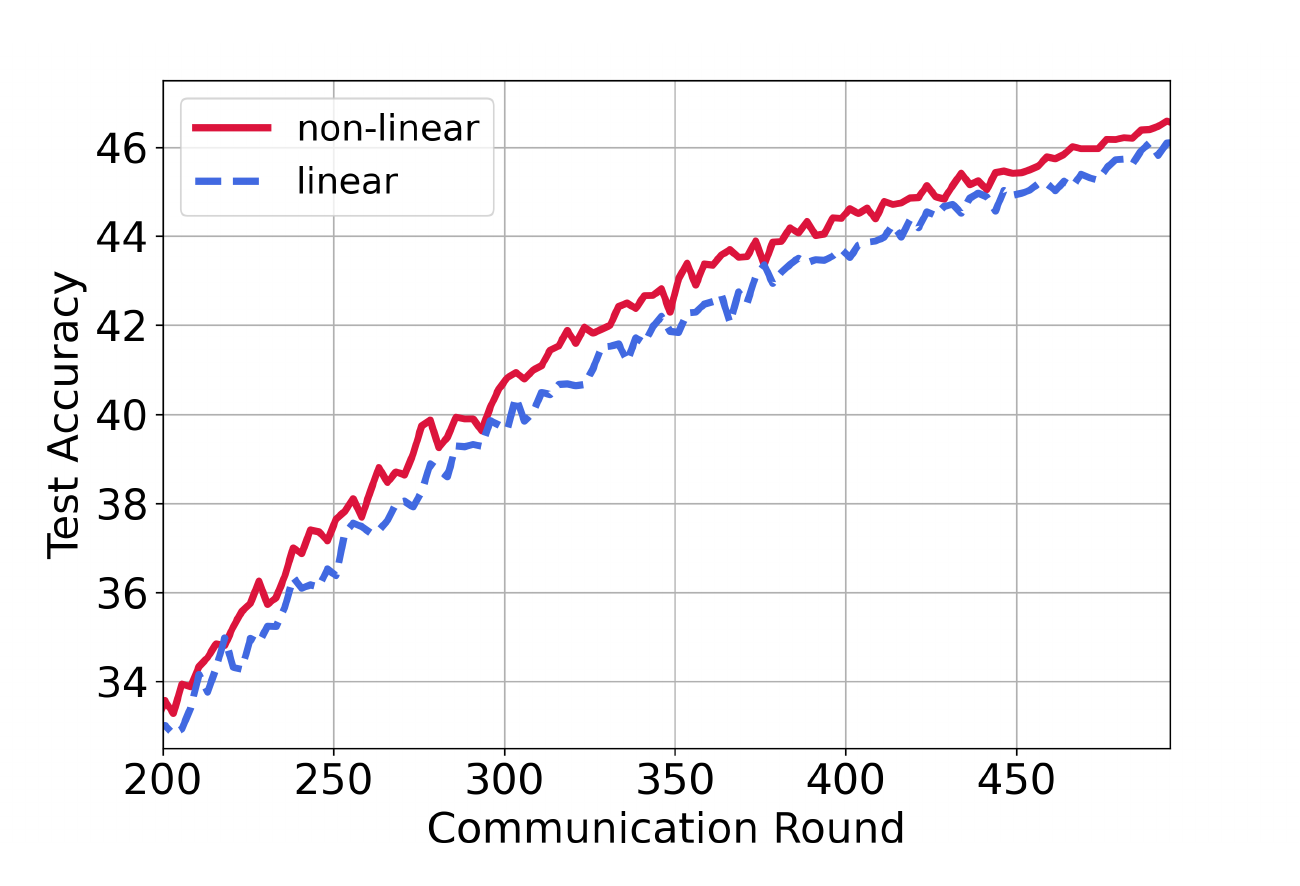} 
\caption{Impact of linear and non-linear $h(\cdot).$}
\label{fig:exp_linear}
\end{figure}

Initially, it is imperative to ascertain the nature (linear vs. non-linear) of the function $h(\cdot)$.
As elucidated in Section~\ref{sec:VAS}, our intuition suggests that the performance degradation due to information aging is non-linear.
To validate it, we compare the test accuracy in both the linear ($h(x) = x$) and nonlinear ($h(x) = \exp(x)$) cases, as illustrated in Figure~\ref{fig:exp_linear}. 
The result reveals a higher test accuracy in the non-linear case. 
Consequently, in all subsequent experiments, we will employ the non-linear function $h(x)$.

\begin{figure}[h]
\centering\includegraphics[width=0.5\textwidth]{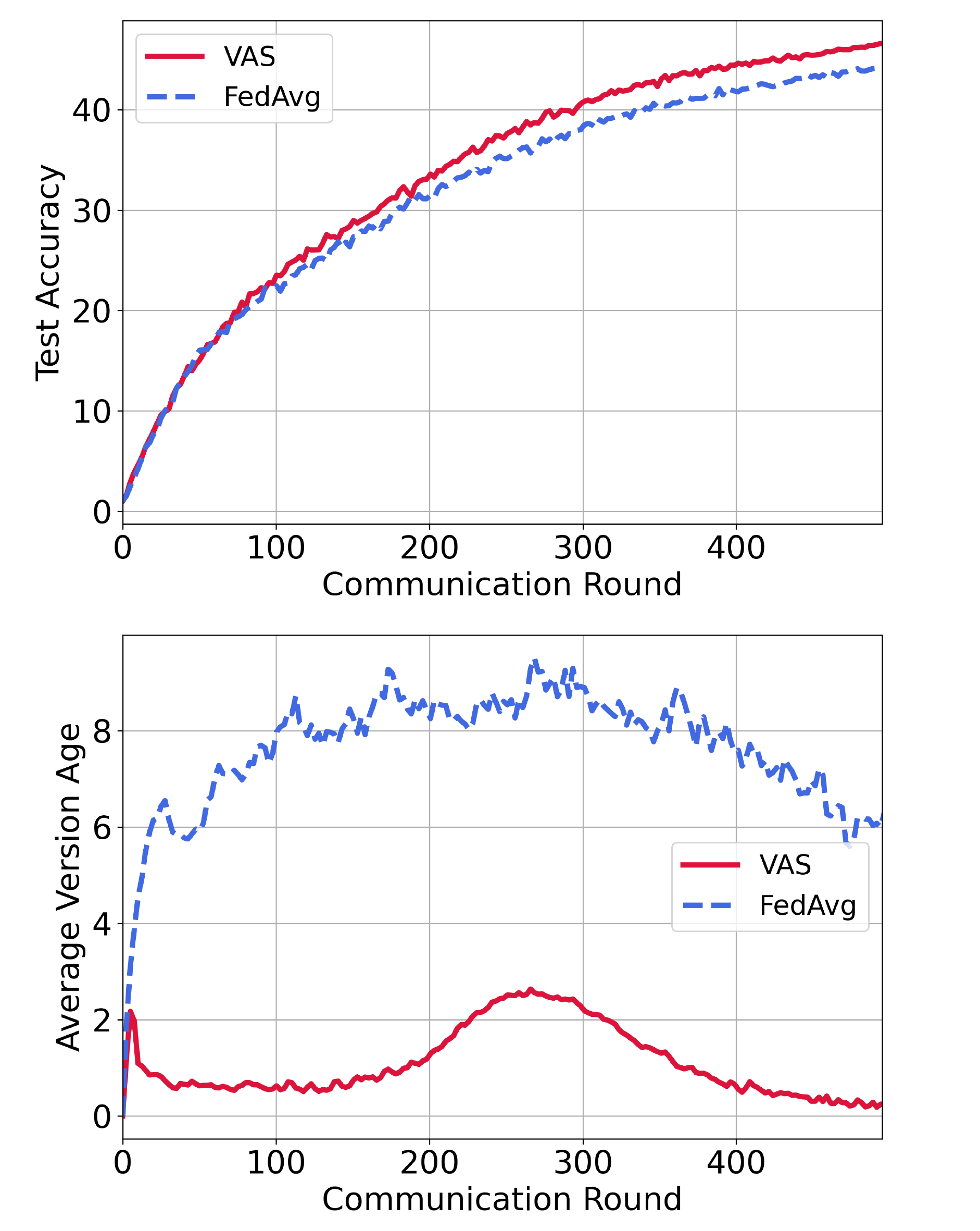} 
\caption{Impact of scheduling policy on the convergence of FL in terms of test accuracy and average version age.}
\label{fig:exp_var_acc}
\end{figure}

Next, we compare the proposed VAS with FedAvg as Figure~\ref{fig:exp_var_acc} depicted.
Notably, the test accuracy value exhibits improvement under the VAS policy compared to FedAvg.
To further explore the factors contributing to the gain, we also visually represented the evolution of the average version age of all clients across communication rounds. 
Throughout the training period, the average version age of the VAS reaches its peak (2.8) at communication round 275 and subsequently decreases gradually towards zero. 
This trend signifies that our method ensures the freshness of system updates. 
In contrast, the average version age of the FedAvg remains above 6.

This divergence arises from FedAvg's practice of randomly selecting clients for updates with equal probability in each communication round. 
An inherent issue with this approach is that certain clients may go unselected for extended periods, causing their local models to progressively diverge from the latest global model. Consequently, the version age continues to increase for these unselected clients. 
Once such clients are eventually chosen, their excessively outdated local updates introduce a substantial deviation to the global model.
In contrast, VAS tends to select clients with larger version age, mitigating the occurrence of outdated local updates. 
Therefore, VAS not only accelerates the convergence rate but also acts as a preventive measure, contributing to enhancing the robustness of the training process.
The aforementioned results indicate that the average version age serves as a valid metric for evaluating the effectiveness and stability of federated systems.

\section{Conclusion}
In this study, we have introduced the concept of version age for FL and investigated the correlation between the average version age and its convergence. Utilizing the version age metric, we devised a scheduling policy for FL that mitigates the impact of excessively outdated local updates, thereby enhancing the overall effectiveness and stability of the system. 

Since this study focused solely on a simple network and dataset, further analysis in more complex scenarios is required. Additionally, the convergence analysis of this scheduling policy requires further investigation.

\bibliographystyle{IEEEbib}
\bibliography{refs}
\end{document}